\documentclass[letterpaper, 10 pt, conference]{ieeeconf}  

\IEEEoverridecommandlockouts                              

\let\labelindent\relax

\usepackage[T1]{fontenc} 
\usepackage{amsthm}
\usepackage{amsmath}
\usepackage{amssymb}
\usepackage{commath}
\usepackage{bm} 
\usepackage{multirow}
\usepackage{enumerate}
\usepackage{enumitem}
\usepackage{wrapfig}
\usepackage{pifont}%
\usepackage{footnote}
\usepackage{threeparttable}
\usepackage{epsfig}
\usepackage{graphicx}
\usepackage{subcaption}
\usepackage{url}            
\usepackage{booktabs}       
\usepackage{amsfonts}       
\usepackage{nicefrac}       
\usepackage{microtype}      
\usepackage{lipsum}

\usepackage{physics}
\usepackage{tabularx}
\usepackage[normalem]{ulem}

\makeatletter
\let\NAT@parse\undefined
\makeatother
\usepackage[colorlinks = true,
            linkcolor = blue,
            urlcolor  = blue,
            citecolor = blue,
            anchorcolor = blue]{hyperref}

\useunder{\uline}{\ul}{}

\newcommand{\bbR}{\mathbb{R}}
\newcommand{\E}{\operatorname{\mathbb{E}}}
\newcommand{\fref}[1]{Fig.~\ref{#1}}
\newcommand{\sref}[1]{Section~\ref{#1}}
\newcommand{\tref}[1]{Table~\ref{#1}}

\title{\LARGE \bf
AirLoop: Lifelong Loop Closure Detection
}

\author{Dasong Gao, Chen Wang$^{\dagger}$, and Sebastian Scherer
\thanks{$^{\dagger}$Corresponding Author. This work was partially sponsored by ONR grant \#N0014-19-1-2266 and ARL DCIST CRA award W911NF-17-2-0181.}
\thanks{The  authors  are  with  the  Robotics  Institute, School of Computer Science,  Carnegie  Mellon  University, Pittsburgh, PA 15213, USA. E-mail: dasongg@andrew.cmu.edu; chenwang@dr.com; basti@andrew.cmu.edu}%
\thanks{Source Code: \url{https://github.com/wang-chen/AirLoop}.}
}

\begin{document}

\maketitle
\thispagestyle{empty}
\pagestyle{empty}

\begin{abstract}
Loop closure detection is an important building block that ensures the accuracy and robustness of simultaneous localization and mapping (SLAM) systems. Due to their generalization ability, CNN-based approaches have received increasing attention. Although they normally benefit from training on datasets that are diverse and reflective of the environments, new environments often emerge after the model is deployed. It is therefore desirable to incorporate the data newly collected during operation for incremental learning. Nevertheless, simply finetuning the model on new data is infeasible since it may cause the model's performance on previously learned data to degrade over time, which is also known as the problem of catastrophic forgetting. In this paper, we present AirLoop, a method that leverages techniques from lifelong learning to minimize forgetting when training loop closure detection models incrementally. We experimentally demonstrate the effectiveness of AirLoop on TartanAir, Nordland, and RobotCar datasets. To the best of our knowledge, AirLoop is one of the first works to achieve lifelong learning of deep loop closure detectors.
\end{abstract}


\section{Introduction}

Loop closure detection (LCD) is an important building block of modern simultaneous localization and mapping (SLAM) systems.
Despite tremendous efforts to improving its accuracy and robustness, visual SLAM is still vulnerable to cumulative errors \cite{zhang2017loop}.
LCD helps combat this problem by identifying the revisited scenes and places,
which allows the robot to reduce localization and mapping drift introduced by sensor measurement errors and scene abnormalities such as occlusion and motion blur \cite{arshad2021role}.

Traditional LCD algorithms leverage handcrafted features like HoG \cite{dalal2005histograms}, SIFT \cite{lowe2004distinctive}, and SURF \cite{bay2008speeded} to generate visual words and build the bag-of-words (BoW) model \cite{nister2006scalable}.
Such methods are, however, vulnerable to environmental changes including illumination, weather, and viewpoint, which can greatly alter the local texture \cite{zhang2017loop}.
In contrast, CNN-based approaches have received increasing attention due to their higher precision, robustness, and generalization ability \cite{masone2021survey}.

However, CNN-based approaches are data-driven.
To maintain the aforementioned advantages, it requires training data to be diverse and reflective of the working environment \cite{tartanair2020iros}.
This cannot be always satisfied
since the new environments we encounter after model deployment may look different from the training set \cite{wang2021unsupervised, garcia2018ibow}.
In this sense, we often expect that a model can learn from new data incrementally.

\begin{figure}[t]
    \includegraphics[width=0.96\linewidth]{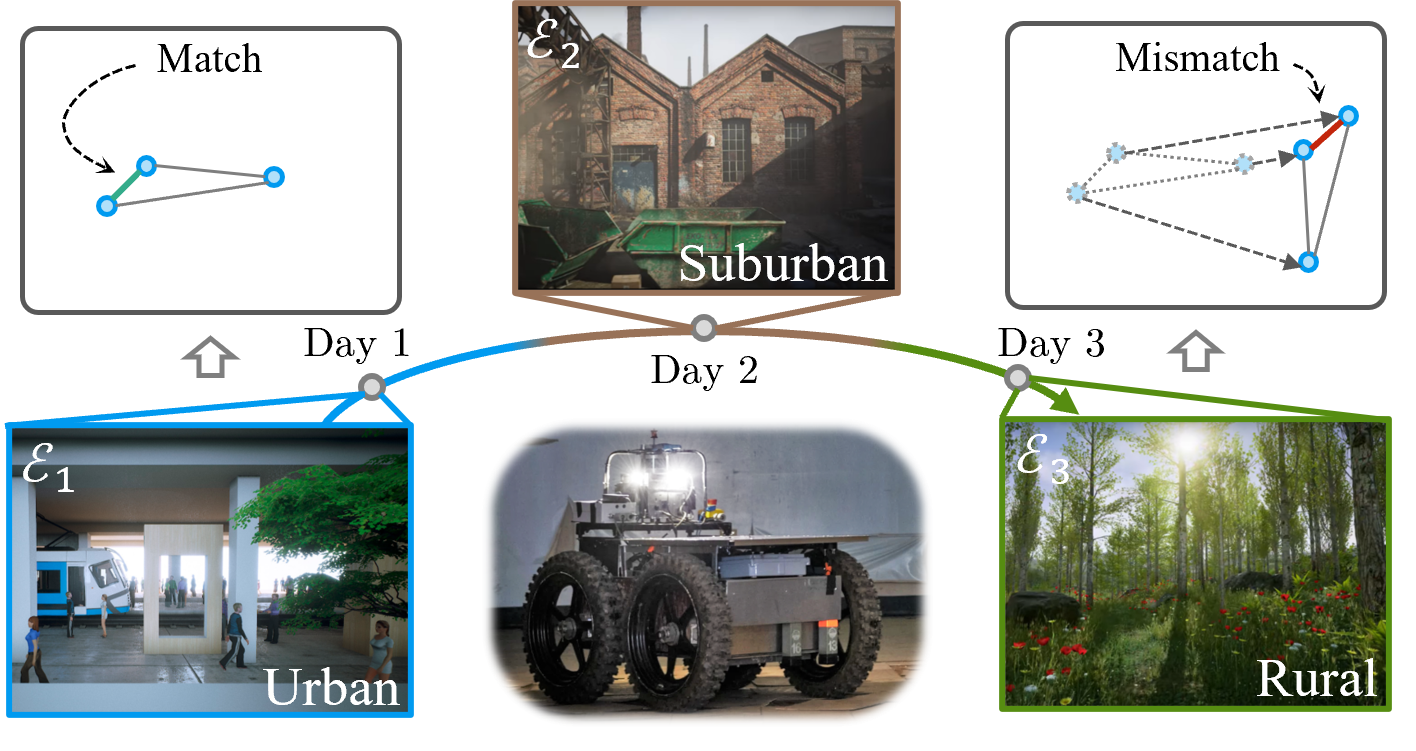}
    \caption{A robot may encounter a series of new environments ($\mathcal{E}_1,\mathcal{E}_2,\dots$) after deployment, in which case the robot may benefit from incremental learning in new environments. However, directly retraining on newly collected data deforms previously learned descriptors, which may cause similar descriptors to separate and distinct descriptors to come close. This will lead to ambiguity and eventually mismatches that are harmful for loop closure detection. In this paper, we resort to lifelong learning to tackle such issues.}
    \label{fig:motivation}
\end{figure}

Nevertheless, continual learning in different environments is difficult.
Unlike traditional methods such as BoW, which learns new scene representation by simply expanding the visual vocabulary with new data \cite{garcia2018ibow}, deep LCD models typically need to adjust all parameters jointly to adapt to new environments.
This poses a dilemma: On the one hand, if we only retrain on the data increments, fitting to the new data almost inevitably alters the parameters and, subsequently, the descriptor space learned from the old data, leading to performance degradation over time, as shown in \fref{fig:motivation}.
This phenomenon is referred to as catastrophic forgetting \cite{lesort2019continual, parisi2019continual}.
On the other hand, if all previously observed data is retained to perform joint training after observing each new environment, the storage cost will grow linearly as the model learns from one environment to the next.
This is prohibitive on resource-constrained devices such as a drone's onboard computer.
The question is therefore: Can we train a deep LCD model from a video stream without suffering from forgetting?

In this work, we address the two problems of incrementally learning a deep LCD model by proposing AirLoop, a lightweight lifelong learning method for loop closure detection.
Specifically, we adapt two lifelong learning techniques that have been used for image classification: memory aware synapses (MAS) \cite{aljundi2018memory} and knowledge distillation (KD) for lifelong embedding learning \cite{hou2018lifelong}.
We also propose a similarity-aware memory buffer that caches samples from a small moving window on the data stream.
This allows us to train the LCD model in a series of environments sequentially without suffering from serious catastrophic forgetting while only requiring constant memory and computation.
To the best of our knowledge, AirLoop is one of the first work to study LCD in the lifelong learning context.

In summary, our contribution includes:
\begin{itemize}
    \item For the first time, we formulate the lifelong loop closure detection problem to study possible ways to continually improve the robustness of a visual SLAM system. 

    \item We introduce a relational variant of MAS for embedding learning in the context of lifelong loop closure detection.

    \item We propose a similarity-aware memory buffer that allows efficient triplet sampling from streaming data.

    \item Through extensive experiments on three datasets, we demonstrate the advantage of our method in alleviating catastrophic forgetting and encouraging generalization.
\end{itemize}

\section{Related Work}


\subsection{Lifelong Learning}
\label{sec:related:lifelong}

Lifelong learning, also known as continual, incremental, or sequential learning, aims at incrementally building up knowledge from an infinite stream of data \cite{delange2021continual}.
The most challenging aspect of lifelong learning is the catastrophic forgetting problem, where fitting to newly arrived data or tasks deteriorates the model's performance on previously observed ones \cite{lesort2019continual, parisi2019continual}.
Recent works on overcoming catastrophic forgetting roughly fall into three families: rehearsal, parameter isolation, and parameter regularization \cite{delange2021continual}.

Rehearsal methods such as iCaRL \cite{rebuffi2017icarl}, GEM \cite{lopez2017gradient}, ER \cite{rolnick2018experience}, and SER \cite{isele2018selective} selectively store a small subset of observed data as exemplars and replay them when new data is introduced.
Some recent methods \cite{shin2017continual, atkinson1802pseudo} replay synthetic samples from the generative adversarial networks (GANs).

Parameter isolation methods alleviate the catastrophic forgetting by dedicating model parameters to specific tasks \cite{delange2021continual}.
Additive approaches such as LwF \cite{li2017learning} and EBLL \cite{rannen2017encoder} introduce new classification heads for each incoming set of classes.
In contrast, subtractive approaches like PackNet \cite{mallya2018packnet} and Piggyback \cite{mallya2018piggyback} incrementally freeze the trained part of the network to achieve theoretically zero forgetting.

Regularization methods take effect by protecting parameters that are important for previously learned tasks.
To define importance, EWC \cite{kirkpatrick2017overcoming} uses the estimated diagonal of the model's Fisher information matrix through back-propagation.
Similarly, SI \cite{zenke2017continual} calculates the per-parameter contribution to loss decrease during the training of previous tasks.
Riemannian Walk \cite{chaudhry2018riemannian} combines the path integral from SI and an online version of EWC.
MAS \cite{aljundi2018memory} defines the importance as the model's sensitivity to the parameters, which is the magnitude of the gradient.
To retain the pairwise embedding distance of previously learned relations, \cite{chen2020exploration} imposes a maximum mean discrepancy (MMD)-based distillation loss.
\cite{chen2021feature} incorporates all previous teacher's knowledge by estimating their outputs with a random perturbation process. To tackle lifelong learning in graph networks, \cite{wang2020lifelong} converted the node classification to graph classification via feature interaction.

Our method includes a variant of MAS (RMAS) optimized for contrastive learning, where we measure the sensitivity of descriptors' pairwise similarities instead of the exact values.
In the experiment section, we show that our relational MAS gives a better performance than the original MAS.
Our work is also related to \cite{chen2021feature} as we also employ a relational knowledge distillation (RKD) loss to further improve its efficacy.
By combining both parameter regularization and knowledge distillation, our method leads to better preservation of previously learned knowledge as shown by the experiments.

\subsection{Loop Closure Detection}

Loop closure detection aims at identifying images that capture the same place of the 3D space from a large image database \cite{masone2021survey}.
To achieve this, traditional LCD methods like 
Galvez-L\'{p}ez and Tard\'{o}s \cite{galvez2012bags}, FAB-MAP \cite{cummins2008fab}, and FAB-MAP 3D \cite{paul2010fab} build a BoW model with visual words generated from handcrafted features offline.
More recently, IBuILD \cite{khan2015ibuild} proposes to dynamically build an environment-specific vocabulary, which is improved in iBoW-LCD \cite{garcia2018ibow} by introducing a tree indexing structure and a RANSAC-based refine step to achieve higher efficiency and accuracy.

In deep learning-based methods, a CNN is trained to generate vector-valued global descriptors for each image, and matching images are retrieved based on descriptor similarities.
Many of such networks employ an extractor-aggregator architecture.
Specifically, given an input image, a feature extractor based on backbones such as VGG \cite{simonyan2014very}, ResNet \cite{he2016deep}, or Inception \cite{szegedy2016rethinking} generates a dense feature map, which is then aggregated into a single vector with methods like NetVLAD \cite{arandjelovic2016netvlad}, R-MAC \cite{gordo2017end}, or GeM \cite{radenovic2018fine}.

Despite having similar goals, our setting differs from the majority of deep LCD in that while regular LCD training allows access to the \textit{full} dataset throughout the training process, data samples in our setting become available only \textit{gradually} in a \textit{sequential order}.
It is, therefore, necessary to take measures against the catastrophic forgetting issue.

We also note that there is one line of work known as long-term LCD, which sounds similar to but is different from lifelong LCD.
Long-term LCD aims at improving the robustness of loop closure detection against environmental changes such as season, weather, and illumination by learning a unified representation of the environment \cite{masone2021survey, zhang2021reference}.
However, to achieve this consistency, long-term LCD models are allowed access to the entire dataset during training \cite{han2018learning, khaliq2019holistic, chen2018learning}, which is not possible under the lifelong setting.
Further, many of the long-term LCD methods employ mechanisms for focusing on elements of the scene that are more stable under environmental changes \cite{linegar2015work, neubert2019neurologically, han2018learning, khaliq2019holistic, chen2018learning, an2020fast}.
However, none of them studied the catastrophic forgetting during knowledge accumulation.
This makes our work the first to consider LCD in the lifelong learning context.

\section{Problem Formulation}
\label{sec:problem}

For completeness, we start with the regular loop closure detection.
Let environment $\mathcal{E} = (\mathcal{I}, l)$, where $\mathcal{I}$ is the set of all images captured from the environment and $l: \mathcal{I} \times \mathcal{I} \mapsto \{0, 1\}$ is the ground truth indicator that outputs 1 for image pairs forming a loop.
Given a query image $I \in \mathcal{I}$ and database images $\mathbf{D} \subset \mathcal{I}$, the regular loop closure detection $C(I, \mathbf{D})$ aims at finding the images in $\mathbf{D}$ that form a loop with $I$:
\begin{equation} \label{eqn:lcd}
    C(I, \mathbf{D}) = \left\{ I' \in \mathbf{D} : l(I', I) = 1\right\}.
\end{equation}
When training on a dataset comprised of $T$ environments, $\{\mathcal{E}^{(i)}\}_{t=1}^T$, where $\mathcal{E}^{(t)} = (\mathcal{I}^{(t)}, l^{(t)})$, the model is allowed to access the full set of images $\bigcup_{t=1}^T\mathcal{I}^{(t)}$ at every training step.
After training, the model is asked to predict loop closure in an observed or unobserved environment:
\begin{equation} \label{eqn:lcdtest}
    C^{(t)}(I, \mathbf{D}) = \left\{ I' \in \mathbf{D} : l^{(t)}(I', I) = 1\right\},
\end{equation}
where $I \in \mathcal{I}^{(t)}, \mathbf{D} \subset \mathcal{I}^{(t)}$.

In lifelong loop closure detection, one is allowed to access the environments and the frames in the environments \textit{only in sequential order}, and each data sample is available \textit{only once}.
In other words, the model learns from the labeled video stream one frame at a time:
\begin{equation}
    \label{eqn:stream}
    (I_1^{(1)}, l^{(1)}), \dots, (I_{N_1}^{(1)}, l^{(1)}), (I_{1}^{(2)}, l^{(2)}), \dots,
\end{equation}
where all $N_i$ images from environment $i$ are observed before switching to the next.
As mentioned earlier, this is due to the limited computational and storage resources of a robot's onboard computer, which forbids joint training using the full set of historical data.
Such a setting is very challenging in that: 1) Since samples are observed sequentially, it is impossible to preprocess the dataset
as required when training most deep loop closure detection models;
2) It is impossible to revisit previously observed environments. In the case where environments look drastically different from each other, sequential training can cause the network to forget knowledge learned from previous environments in favor of learning a new environment, which is known as the catastrophic forgetting.

\section{Method}
\label{sec:method}

\subsection{Contrastive Learning for Loop Closure Detection}
\label{sec:contrastive}

We follow the common visual place recognition pipeline \cite{masone2021survey} by learning an LCD model for generating global descriptors and using descriptor similarities to predict loop closure.
Concretely, let $\mathcal{I} = \bigcup_{i=1}^T \mathcal{I}^{(t)}$ be the collection of all images from all environments and $f(\cdot; \theta): \mathcal{I} \mapsto \bbR^{D}$ be the descriptor network parametrized by $\theta$ that produces a global descriptor of length $D$ for a given input image.
Let $I \in \mathcal{I}^{(t)}$ and $\mathbf{D} \subset \mathcal{I}^{(t)}$ be the query and database from environment $t$, we retrieve the subset of $\mathbf{D}$ with sufficiently high descriptor similarity:
\begin{equation} \label{eqn:lcd-deep}
    \widetilde{C}^{(t)}(I, \mathbf{D}) = \left\{ I' \in \mathbf{D} : \operatorname{sim}(f(I'), f(I)) \geq 1 - \epsilon \right\}.
\end{equation}
where $\epsilon\in (0, 1)$ is a constant and $\operatorname{sim}(\cdot, \cdot)$ is the cosine similarity.
Let $I_a, I_p, I_n \in \mathcal{I}^{(t)}$ be a triplet of images (namely, the anchor, the positive, and the negative) from the $t$-th environment, such that for the positive (loop-closing) pair, $l^{(t)}(I_a, I_p) = 1$ and for the negative (irrelevant) pair, $l^{(t)}(I_a, I_n) = 0$.
We calculate descriptor similarities $s_{ap} = \operatorname{sim}(f(I_a), f(I_p))$ and $s_{an} = \operatorname{sim}(f(I_a), f(I_n))$ and apply the triplet loss \cite{schroff2015facenet}, which can be represented as
\begin{equation} \label{eqn:triplet-loss}
    L_{\text{triplet}}^{(t)} = \max(s_{an} - s_{ap} + \delta, 0),
\end{equation}
where $\delta$ is a constant margin.
Intuitively, this forces the network to predict similar descriptors for the positive pair $(I_a, I_p)$ but distinct descriptors for the negative pair $(I_a, I_n)$.
However, due to the sequential data access constraint, it is difficult to sample triplets from the data stream \eqref{eqn:stream}.
We next demonstrate a triplet sampling technique for streaming data.

\subsection{Buffered Triplet Sampling}
\label{sec:memory}

In regular contrastive learning, the triplet is usually uniformly sampled from the set of all possible triplets of the current environment: $(I_a, I_p, I_n) \sim \{ (I'_a, I'_p, I'_n) \in \mathcal{I}^{(t)}: l^{(t)}(I'_a, I'_p) = 1, l^{(t)}(I'_a, I'_n) = 0\}$.
However, given constraints in \sref{sec:problem} that the data items are only available sequentially, this scheme is not directly applicable.
To resolve the issue, we maintain a similarity-aware memory buffer $(\mathcal{M}^{(t)}, \mathbf{S}^{(t)})$ for environment $t$ to temporarily store a fixed number of past images together with the ground-truth from $l^{(t)}$.
Specifically, the memory $\mathcal{M}^{(t)}$ is a first-in-first-out image queue of size $M$ and the matrix $\mathbf{S}^{(t)} \in \{0, 1\}^{M \times M}$ records the pairwise relationship for images in $\mathcal{M}^{(t)}$ such that
\begin{equation}
    \label{eqn:similarity}
    S_{ij}^{(t)} = l^{(t)}(\mathcal{M}^{(t)}[i], \mathcal{M}^{(t)}[j]).
\end{equation}
When a new image arrives, we replace the oldest image in $\mathcal{M}^{(t)}$ with the new image and update $\mathbf{S}^{(t)}$ according to \eqref{eqn:similarity}.

To sample a triplet from $\mathcal{M}^{(t)}$, we first uniformly randomly select an anchor $I_a = \mathcal{M}^{(t)}[a]$.
The positive $I_p$ and negative $I_n$ are then uniformly randomly drawn from the buffer such that $S_{an} = 0$ and $S_{ap} = 1$.

\subsection{Relational Memory Aware Synapses}
\label{sec:rmas}
It is commonly hypothesized in regularization-based lifelong learning methods that catastrophic forgetting happens when learning of new tasks alters the network parameters that are important for previously learned tasks \cite{lesort2019continual, masana2020class}.
To prevent such undesirable updates when learning the $t$-th task, MAS \cite{aljundi2018memory} assigns an importance weight ${}^{\text{MAS}}\Omega_i^{(t)}$ to each parameter $\theta_i$
and penalize its changes by the regularization loss.
In this paper, we use a relational MAS (RMAS) loss:
\begin{equation}
    \label{eqn:importance}
    L_{\text{RMAS}}^{(t)} = \sum_{i = 1}^m {}^{\text{RMAS}}\Omega_i^{(t - 1)} \left(\theta_i - \theta_i^{(t - 1)}\right)^2,
\end{equation}
where $m$ is the number of network parameters and $\theta_i^{(t - 1)}$ is the parameter after learning tasks $1, \dots, t-1$.

In MAS \cite{aljundi2018memory}, $^{\text{MAS}}\Omega_i^{(t)}$ is estimated by the sensitivity of the network's output to the corresponding parameter:
\begin{equation}
    \label{eqn:mas}
    ^{\text{MAS}}\Omega_i^{(t)} = \E_{I \sim \mathcal{I}^{(t)}} \left(\pdv{\norm{f(I)}_2}{\theta_i}\right)^2,
\end{equation}
where $I$ is drawn from the image stream of the $t$-th environment and $\norm{\cdot}_2$ is the $\ell_2$-norm.
Unlike in classification tasks, where the components of network output are tied to fixed classes, LCD pays less attention to the exact values of place descriptors but focuses more on their pairwise similarities.
Therefore, in RMAS, we instead penalize the changes to the pairwise similarity $\operatorname{sim}(f(I), f(I'))$ by setting
\begin{equation}
    \label{eqn:rmas}
    ^{\text{RMAS}}\Omega_i^{(t)} = \E_{I, I' \sim \mathcal{I}^{(t)}} \left(\pdv{\operatorname{sim}(f(I), f(I'))}{\theta_i}\right)^2.
\end{equation}
Directly evaluating \eqref{eqn:rmas} is difficult under the lifelong learning setting since it requires the availability of all sample pairs from the current environment.
Moreover, in an environment with $N$ images, the derivative in \eqref{eqn:rmas} is calculated $\nicefrac{N(N-1)}{2}$ times.
To make it computationally tractable, we approximate it by accumulating the squared gradient of the Gram matrix's Frobenius norm from triplets sampled at each training step:
\begin{equation}
    \label{eqn:rmas-approx}
    ^{\text{RMAS}}\Omega_i^{(t)} \approx \frac{1}{N_t} \sum_{k=1}^{N_t} \left(\pdv{\|\mathbf{\widetilde{S}}_k^{(t)}\|_F}{\theta_i}\right)^2,
\end{equation}
where $\norm{\cdot}_F$ is the Frobenius norm
and $\mathbf{\widetilde{S}}_k^{(t)} \in \bbR^{3 \times 3}$ is the Gram matrix at the $k$-th training step in environment $t$, which is formed by the output triplet $f(I_a), f(I_p), f(I_n)$ such that $\widetilde{S}_{k, 12} = \operatorname{sim}(f(I_a), f(I_p)), \widetilde{S}_{k, 13} = \operatorname{sim}(f(I_a), f(I_n))$, etc.
After collecting importance weights $^\text{RMAS}\Omega_i^{(t-1)}$ in the $\mathcal{E}^{(t-1)}$, we use \eqref{eqn:importance} to protect the important parameters in $\mathcal{E}^{(t)}$.

\subsection{Relational Knowledge Distillation}
\label{sec:rkd}

It is known that knowledge distillation can be combined with parameter regularization to achieve less forgetting \cite{peng2021defense}.
Therefore, we apply the knowledge distillation loss to alleviate forgetting by forcing the current model $f(I, \theta)$ to retain the descriptor distances learned in environment $1, \dots, t - 1$, which we call relational knowledge distillation (RKD):
\begin{equation}
    \label{eqn:rkd}
     \begin{split}
         L_{\text{RKD}}^{(t)}
         &= \|\mathbf{\widetilde{S}}^{(t)} - \mathbf{\widetilde{S}}^{(t - 1)}\|_F,
     \end{split}
\end{equation}
where $\mathbf{\widetilde{S}}^{(t)}$ is the triplet Gram matrix for the current training step defined in \eqref{eqn:rmas-approx},
and $\mathbf{\widetilde{S}}^{(t - 1)}$ is its counterpart produced by $f(I, \theta^{(t - 1)})$.
Note that the input images $I_a, I_p, I_n$ are drawn from the current environment, eliminating the need to preserve data from previous environments.
This approach is similar to \cite{chen2021feature}, where $\mathbf{\widetilde{S}}^{(t - 2)}, \dots, \mathbf{\widetilde{S}}^{(1)}$ are additionally estimated and used as targets in  \eqref{eqn:rkd}.
However, we experimentally found that it is sufficient to only consider \eqref{eqn:rkd} in our setting.

\subsection{Combined Loss for Lifelong Learning}
\label{sec:rmasrkd}

We adopt a combined loss $L^{(t)}$ for the $t$-th environment: 
\begin{equation}
    \label{eqn:total}
    L^{(t)} = L^{(t)}_{\text{triplet}} + \lambda_1 L^{(t)}_{\text{RMAS}} + \lambda_2 L^{(t)}_{\text{RKD}},
\end{equation}
where $\lambda_1, \lambda_2$ are hyperparameters.
In the experiments, we show that although both $L^{(t)}_{\text{RMAS}}$ and $L^{(t)}_{\text{RKD}}$ alone is effective at alleviating forgetting and encouraging generalization, the combined loss yields a noticeably better result.

\begin{table*}[ht]
    \centering
    \begin{threeparttable}
        \caption{Performance comparison for lifelong loop closure detection. $^\dagger$}
        \label{tab:lifelong}
        \setlength{\tabcolsep}{2.5pt}
        \begin{tabularx}{\linewidth}{Xccccccccc}
\toprule
\multicolumn{1}{l}{} & \multicolumn{3}{c}{TartanAir}                                                  & \multicolumn{3}{c}{Nordland}                                                   & \multicolumn{3}{c}{RobotCar}                                                   \\
\cmidrule(lr){2-4} \cmidrule(lr){5-7} \cmidrule(lr){8-10}
\multicolumn{1}{c}{} & \multicolumn{1}{c}{AP} & \multicolumn{1}{c}{BWT} & \multicolumn{1}{c}{FWT} & \multicolumn{1}{c}{AP} & \multicolumn{1}{c}{BWT} & \multicolumn{1}{c}{FWT} & \multicolumn{1}{c}{AP} & \multicolumn{1}{c}{BWT} & \multicolumn{1}{c}{FWT} \\
\midrule
Finetune                  & 0.754$\pm$0.002                & -0.009$\pm$0.006            & 0.730$\pm$0.001              & 0.615$\pm$0.014                & -0.012$\pm$0.012            & 0.549$\pm$0.012              & 0.411$\pm$0.016                & -0.066$\pm$0.007            & 0.462$\pm$0.011             \\
EWC   \cite{kirkpatrick2017overcoming}    & 0.758$\pm$0.006                & -0.005$\pm$0.003            & 0.728$\pm$0.004              & 0.614$\pm$0.020                & -0.014$\pm$0.010            & 0.549$\pm$0.015        & 0.416$\pm$0.004                & -0.054$\pm$0.017            & 0.461$\pm$0.013             \\
SI  \cite{zenke2017continual}   & 0.753$\pm$0.003                & -0.010$\pm$0.003            & 0.730$\pm$0.002              & 0.614$\pm$0.015                & -0.010$\pm$0.012            & 0.549$\pm$0.012              & 0.407$\pm$0.013                & -0.062$\pm$0.023            & 0.454$\pm$0.005             \\
\midrule
AirLoop$^{\text{w/o RMAS}}$        & 0.757$\pm$0.001                & -0.008$\pm$0.004            & {\ul 0.733$\pm$0.002}              & \textbf{0.632$\pm$0.020}          & {\ul 0.008$\pm$0.009}       & \textbf{0.549$\pm$0.024}              & 0.447$\pm$0.017                & -0.041$\pm$0.027            & 0.472$\pm$0.015             \\
AirLoop$^{\text{w/o RKD}}$     & {\ul 0.759$\pm$0.005}          & {\ul -0.001$\pm$0.002}      & 0.732$\pm$0.001              & 0.622$\pm$0.012                & 0.006$\pm$0.012             & 0.545$\pm$0.013              & {\ul 0.456$\pm$0.006}          & \textbf{-0.010$\pm$0.007}   & \textbf{0.486$\pm$0.007}    \\
AirLoop               & \textbf{0.769$\pm$0.002}       & \textbf{0.007$\pm$0.002}    & \textbf{0.736$\pm$0.003}     & {\ul 0.631$\pm$0.012}                & \textbf{0.018$\pm$0.009}    & {\ul 0.546$\pm$0.017}              & \textbf{0.461$\pm$0.009}       & {\ul -0.013$\pm$0.011}      & {\ul 0.485$\pm$0.013}       \\
\midrule
IFGIR$^*$              & 0.753$\pm$0.003                & -0.008$\pm$0.003            & {0.733$\pm$0.002}        & {0.639$\pm$0.019}       & 0.008$\pm$0.007             & {0.560$\pm$0.020}     & 0.446$\pm$0.007                & -0.030$\pm$0.007            & 0.476$\pm$0.009             \\
Joint$^\ddagger$                 & 0.772$\pm$0.003                & -                       & -                        & 0.650$\pm$0.023                & -                       & -                        & 0.485$\pm$0.011                & -                       & -                      \\
\bottomrule
\end{tabularx}
        \begin{tablenotes}[normal,flushleft]
            \item $^\dagger$We highlight the best performance in each metric with \textbf{bold} and the second-best with \underline{underline}. Each experiment is repeated 3 times. AP (average performance) is measured by recall @ 100\% precision. Forward transfer (FWT) and backward transfer (BWT) are defined in \eqref{eqn:metric}.
            \item $^*$IFGIR \cite{chen2021feature} requires a separate validation dataset for estimating previous outputs from the most recent teacher model, which cannot meet the requirements of lifelong learning, we instead save all historical teacher models during training. This produces a theoretical upper bound performance for IFGIR.
            \item $^\ddagger$Joint training is expected to produce the theoretical upper-bound performance for lifelong learning, thus only average performance can be reported. The performance of AirLoop is close to this upper bound, indicating its effectiveness.
        \end{tablenotes}
    \end{threeparttable}
    \vspace{-6pt}
\end{table*}

\section{Experiments}

\subsection{Network Structure \& Implementation Details}
\label{sec:impl}

We adopt a pretrained VGG-19 as the feature extractor and use GeM \cite{radenovic2018fine} as the feature aggregator to produce a descriptor of dimension 1024.
That is, after obtaining the highest-level feature map, we calculate the mean feature over all pixels and transform it with a two-layer perceptron.
We use the SGD optimizer with a learning rate of 0.002 and a momentum of 0.9.
For each dataset described in \sref{sec:dataset}, we sequentially train the network with our method and the baselines in different environments and use the parameters obtained from environment $t$ to initialize the network at environment $t+1$.
The memory size $M$ is fixed at 1000.

\begin{figure}[t]
    \centering
    \includegraphics[width=0.96\linewidth]{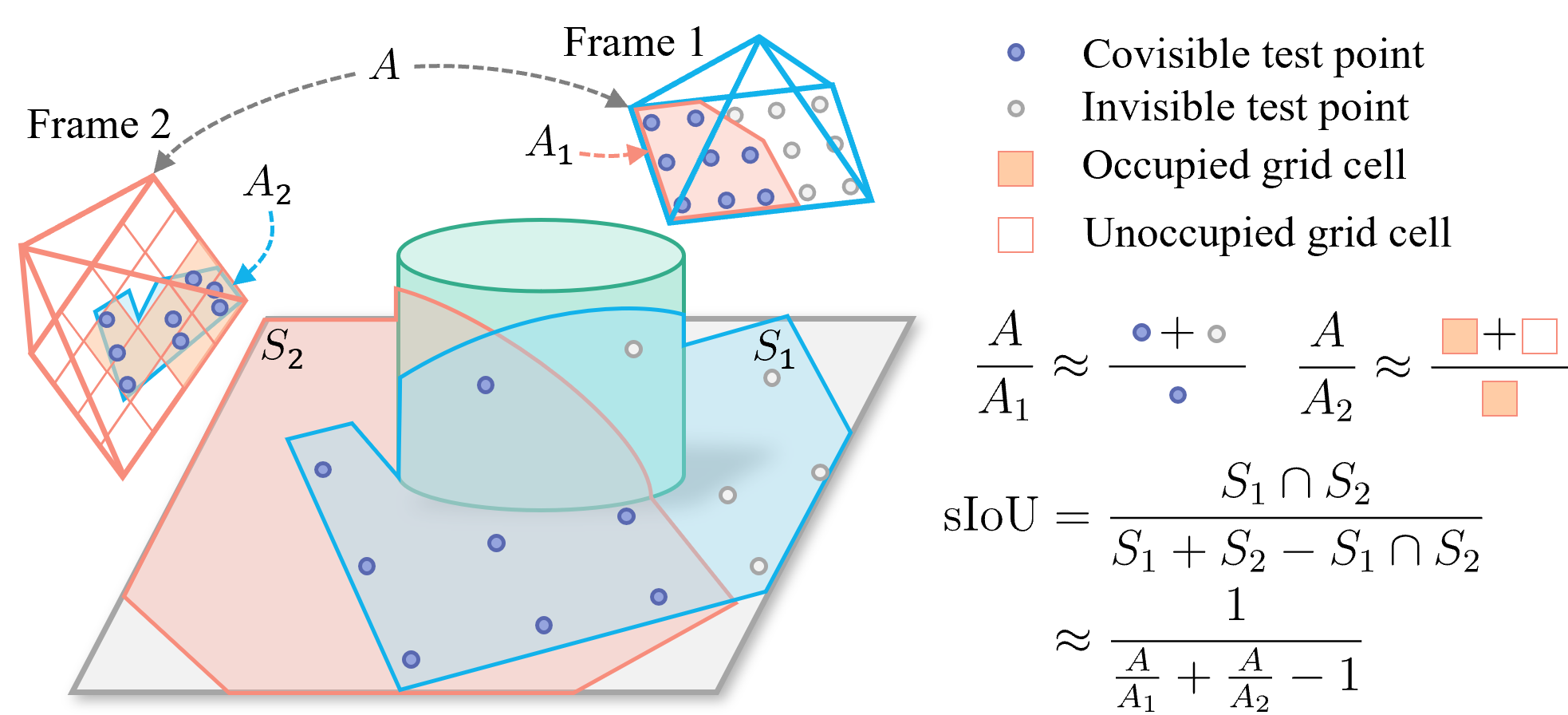}
    \caption{We take the surface IoU to indicate a true loop closure. Specifically, we quantify the overlap between two camera frames by the IoU of surfaces ($S_1$ and $S_2$) captured by both cameras. To make it computationally tractable, we further estimate the area ratios on the screen: $\frac{S_1}{S_1 \cap S_2} \approx \frac{A}{A_1}$ and $\frac{S_2}{S_1 \cap S_2} \approx \frac{A}{A_2}$, which is approximated by projecting a grid of test points from one camera to the other and counting the number of grid cells occupied by the covisible test points.}
    \label{fig:siou}
\end{figure}

\subsection{Datasets}
\label{sec:dataset}

We evaluate our method on one large-scale synthetic dataset, TartanAir \cite{tartanair2020iros} as well as two real-world datasets, Nordland \cite{olid2018nordland}, and RobotCar \cite{RobotCarDatasetIJRR}.

\textbf{Nordland} There are four seasonal environments in the Nordland dataset, namely, spring, summer, fall, and winter.
We use the recommended train-test split and label the pairs with a maximum distance of 3 frames as loop closure.

\textbf{RobotCar} There are three environments based on the lighting condition, labeled sun, overcast, and night.
For each environment, we select two sequences as the training and test set, respectively.
Loop closure is defined as frames with distance less than 10m and yaw difference less than $15^\circ$.

\textbf{TartanAir} 
TartanAir is a large (about 3TB) and very challenging visual SLAM dataset consisting of binocular RGB-D video sequences together with additional per-frame information such as camera poses, optical flow, and semantic annotations.
It features environments with various themes including urban, rural, natural, domestic, sci-fi, etc.
We choose five scenes from the dataset and take 80\% of the sequences from each scene for training and the rest for testing.
Unlike Nordland or RobotCar, whose trajectories span a large space, TartanAir's trajectories are more compact and convoluted.

In some of the sequences, we find that the Euclidean or timestamp distance is not a good indicator of loop closure since two frames taken from spatially close positions may look at different directions and vice versa.
Therefore, we use ``visual overlap'' to define loop closure:
Let $A$ be the screen area and $A_1$, $A_2$ be the area of covisible parts of the scene in the two frames.
We define the surface intersection-over-union (sIoU) in \eqref{eqn:sIoU}, as illustrated in \fref{fig:siou}.
\begin{equation}
    \label{eqn:sIoU}
    \text{sIoU} = \frac{1}{\nicefrac{A}{A_1} + \nicefrac{A}{A_2} - 1}.
\end{equation}
We take $\text{sIoU} > 0.7$ as positive and $\text{sIoU} < 0.1$ as negative during training and take $\text{sIoU} > 0.5$ as positive during testing.

\subsection{Evaluation Metrics}

Precision is paramount for LCD tasks since any incorrect matches may lead to fatal map degradation \cite{garg2021your}.
Therefore, we choose the recall rate at 100\% precision as the metric of model performance.
To better measure the interference between the learning of different environments, we evaluate the model performance on all environments, including the ones that haven't been trained on, after learning each environment. 
Therefore, for a dataset with $T$ environments, we obtain a $T \times T$ performance matrix $\mathbf{R}$, where $R_{i, j}$ denotes the performance on environment $j$ after learning environment $i$.
We adopt the evaluation protocol \eqref{eqn:metric} to summarize $\mathbf{R}$ into three scalar metrics, which were introduced in  \cite{lopez2017gradient}.
\begin{equation}
    \label{eqn:metric}
    \begin{split}
        \text{AP} &= \nicefrac{\left(\sum_{i = 1}^T \sum_{j = 1}^i R_{i, j}\right)}{(T (T + 1) / 2)}, \\
        \text{BWT} &= \nicefrac{\left(\sum_{i = 2}^T \sum_{j = 1}^{i - 1} R_{i, j} - R_{j, j}\right)}{(T (T - 1) / 2)}, \\
        \text{FWT} &= \nicefrac{\left(\sum_{i = 1}^T \sum_{j = i + 1}^{T} R_{i, j}\right)}{(T (T - 1) / 2)}, \\
    \end{split}
\end{equation}
where AP (average performance) is the overall performance on seen environments, BWT (backward transfer) accounts for the influence of future learning on knowledge learned from previous environments, and FWT (forward transfer) measures the generalization in unseen environments.

\subsection{Methods for Comparison}

We compare the following methods:

\subsubsection{Finetune} The network is finetuned on new environments only with the triplet loss $L_{\text{triplet}}$.

\subsubsection{EWC/SI} Two generic regularization-based lifelong learning methods, elastic weight consolidation (EWC) \cite{kirkpatrick2017overcoming} and synaptic intelligence \cite{zenke2017continual}.

\subsubsection{AirLoop} Our method presented in \sref{sec:method}.

\subsubsection{AirLoop$^{\text{w/o RMAS}}$} Our AirLoop network without the RMAS loss presented in \sref{sec:rmas}.

\subsubsection{AirLoop$^{\text{w/o RKD}}$} Our AirLoop network without the RKD loss described in \sref{sec:rkd}.

\subsubsection{IFGIR$^*$} The upper-bound of incremental fine-grained image retrieval (IFGIR) \cite{chen2021feature}. The original method requires a separate validation set, which is not available in the lifelong data setting. Instead, we implement its reported upper-bound of \cite{chen2021feature}, which uses models learned from all previous environments to perform relational knowledge distillation.

\subsubsection{Joint} Triplets can be sampled from any environment in arbitrary order. This gives the performance upper-bound.

\subsection{Performance}
\label{sec:perf}

We report the overall performance in \tref{tab:lifelong}.
It can be seen that the Finetuned model exhibits a more negative negative BWT, which means that forgetting is problematic if lifelong learning techniques are not applied.
In contrast, AirLoop is the only method that exhibits positive BWT on both TartanAir and Nordland, demonstrating its strong ability to not only retain but also extend previously learned knowledge.

In terms of AP / FWT, we only obtain very modest performance from generic regularization-based methods (EWC and SI) despite careful hyperparameter search.
In comparison, AirLoop is able to outperform them by a noticeable margin and is advantageous than IFGIR$^*$ on TartanAir and RobotCar.
This shows the advantage of our methods at encouraging generalization.
We notice that IFGIR$^*$ is dominating in AP and FWT on Nordland.
This is because the scene appearance in Nordland is quite similar for spring, summer, and fall environments, which allows the teacher models to remain informative and facilitates knowledge transfer.
However, one shall not expect environments to be similar in all cases, especially in the setting of lifelong learning.

\begin{figure*}[ht]
    \centering
    \includegraphics[width=0.99\linewidth]{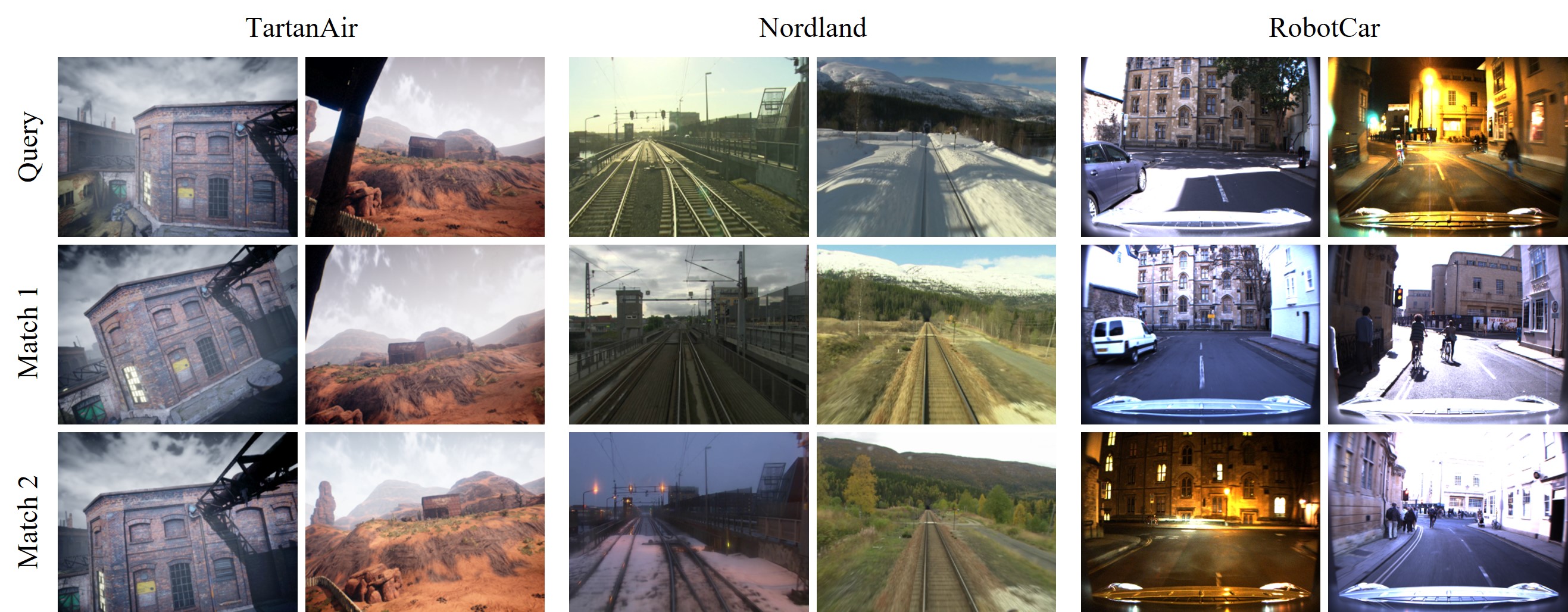}
    \caption{Examples of loop closure detection after learning all environments on each dataset. Note that our model is trained incrementally with triplets from individual environments but is able to perform cross-environment loop closure detection. }
    \label{fig:qualitative}
\end{figure*}

\subsection{Computational Efficiency}

We list the running time of different operations taken during one training step in \tref{tab:runtime-operations}.
It can be observed that storing and sampling from the memory buffer only introduce a small overhead compared with the back-propagation.
We further compare the running time of the forward pass of different lifelong methods in \tref{tab:runtime-methods}.
Note that our method only requires 60\% running time of IFGIR$^*$.

\begin{table}[t]
    \caption{Runtime Analysis.}
    \centering
    \label{tab:runtime}
    \begin{subtable}{.45\linewidth}
      \centering
        \setlength{\tabcolsep}{3.8pt}
        \caption{Runtime of steps}
        \label{tab:runtime-operations}
        \begin{tabularx}{\linewidth}{Xc}
            \toprule
            Operations & Time (ms) \\
            \midrule
            Memory store & 66.5 \\
            Memory sample & 23.4 \\
            Evaluate $L_\text{triplet}$ & 17.8 \\
            Back-propagation & 185.5 \\
            \bottomrule
        \end{tabularx}
    \end{subtable}
    \hspace{10pt}
    \begin{subtable}{.45\linewidth}
      \centering
        \caption{Runtime of methods}
        \label{tab:runtime-methods}
        \begin{tabularx}{\linewidth}{Xc}
            \toprule
            Methods & Time (ms) \\
            \midrule
            IFGIR$^*$ & 452.9 \\
            AirLoop$^{\text{w/o RKD}}$ & 193.1 \\
            AirLoop$^{\text{w/o RMAS}}$ & 97.9 \\
            AirLoop & 292.3 \\
            \bottomrule
        \end{tabularx}
    \end{subtable} 
\end{table}

\section{Ablation Studies}
\label{sec:ablation}

\begin{figure}[b]
    \subfloat[RobotCar]{
        \includegraphics[height=0.44\linewidth]{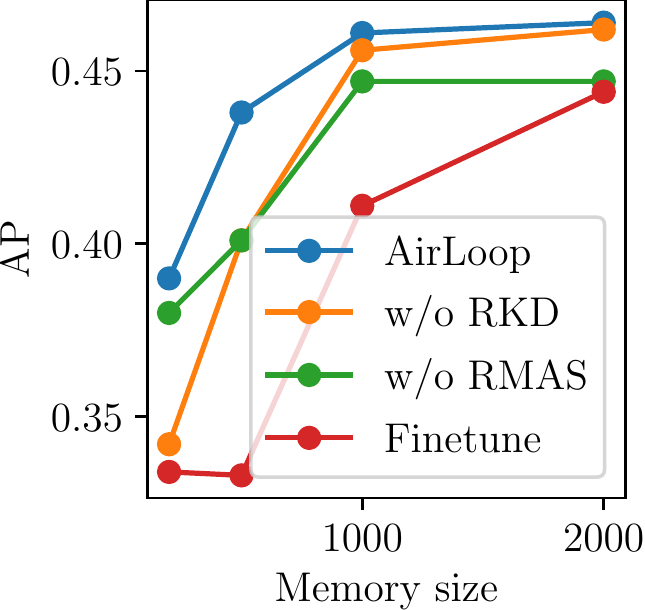}}
    \subfloat[Nordland]{
        \includegraphics[height=0.44\linewidth]{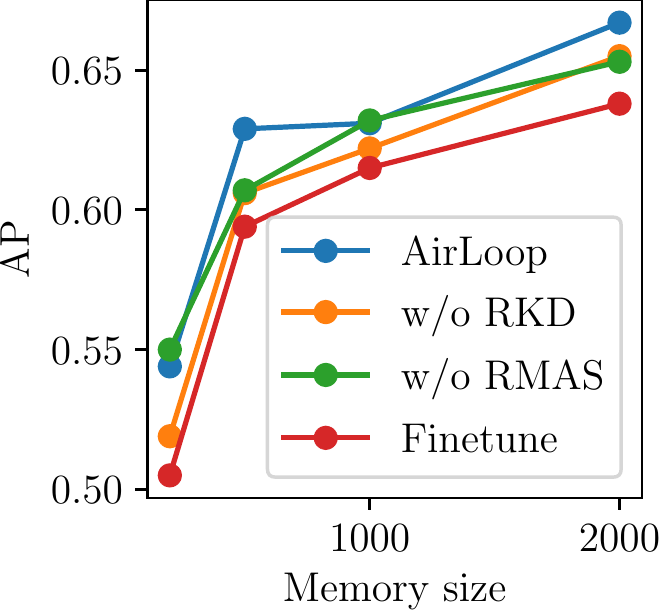}}
    \caption{Impact of memory buffer size on average performance (recall @ 100\% precision) after learning all environments.}
    \label{fig:memory}
\end{figure}

\subsection{Memory Size}

One major function of the memory buffer is to provide more choices of triplets so that the distribution of the sampled triplets can approach that of the whole dataset.
Here we investigate the influence of memory size $M$ on performance.
We take RobotCar and Nordland as examples and present the final AP in \fref{fig:memory}.
It can be seen that with a manageable memory size of a few hundred to 1000, our proposed method is able to reach its peak performance.

\subsection{Relational Knowledge Preservation}

We present the comparison between relational and non-relational variants of MAS and KD on RobotCar in \tref{tab:relational}.
It is observed that our two relational losses (RKD and RMAS) bring higher performance than non-relational losses (KD and MAS), which indicates the effectiveness of our method.
Note that the non-relational losses tend to preserve the absolute values of the original descriptors, while our relational losses only retain the relative relationship between the descriptors.
This allows our methods to focus on maintaining descriptor similarity, the property we care the most about.

\begin{table}[t]
    \centering
    \begin{threeparttable}
        \caption{Comparison between relational and non-relational variant of MAS and KD.}
        \label{tab:relational}
        \setlength{\tabcolsep}{4pt}
        \begin{tabularx}{0.9\linewidth}{Xccc}
            \toprule
            & AP & BWT & FWT \\
            \midrule
            Finetune  & 0.411 $\pm$ 0.016 & -0.066 $\pm$ 0.007  & 0.462 $\pm$ 0.011  \\
            \midrule
            KD  & 0.414 $\pm$ 0.003 & -0.058 $\pm$ 0.014  & 0.467 $\pm$ 0.014  \\
            \textbf{RKD}  & \textbf{0.447 $\pm$ 0.017} & \textbf{-0.041 $\pm$ 0.027}  & \textbf{0.472 $\pm$ 0.015}  \\
            \midrule
            MAS  & 0.446 $\pm$ 0.010 & -0.023 $\pm$ 0.012  & 0.482 $\pm$ 0.010  \\
            \textbf{RMAS}  & \textbf{0.456 $\pm$ 0.006} & \textbf{-0.010 $\pm$ 0.007}  & \textbf{0.486 $\pm$ 0.007}  \\
            \bottomrule
        \end{tabularx}
    \end{threeparttable}
\end{table}

\section{Conclusions}

We propose AirLoop, a method for incrementally training deep loop closure detection models.
To enable triplet sampling, we implement a similarity-aware memory buffer to cache recently observed frames together with their pairwise similarity.
We also formulate two lifelong relation learning losses, RMAS and RKD, for alleviating catastrophic forgetting in loop closure detection.
Extensive experiments demonstrate the advantage of AirLoop in lifelong loop closure detection.
Future work may consider more complicated backbone networks or incorporate memory replay methods.








{
    \bibliographystyle{IEEEtran}
    \bibliography{main}
}

\end{document}